\def\year{2022}\relax
\documentclass[letterpaper]{article} 
\usepackage{aaai22}  
\usepackage{times}  
\usepackage{helvet}  
\usepackage{courier}  
\usepackage[hyphens]{url}  
\usepackage{graphicx} 
\usepackage{natbib}  
\usepackage{caption} 
\DeclareCaptionStyle{ruled}{labelfont=normalfont,labelsep=colon,strut=off} 
\usepackage{algorithm}
\usepackage{algorithmic}

\usepackage{newfloat}
\usepackage{listings}
\floatstyle{ruled}
\newfloat{listing}{tb}{lst}{}
\floatname{listing}{Listing}
\usepackage{bm}
\usepackage{times}
\usepackage{latexsym}
\usepackage{amsmath,amsfonts,amssymb,bbm,epsfig,bm}
\usepackage{booktabs}
\usepackage{multirow}
\usepackage{array}
\usepackage{url}
\usepackage{tikz}
\usepackage{makecell}
\usepackage{subfigure}
\usepackage{framed}
\usepackage{pstricks}
\usepackage{xcolor}
\usepackage{enumerate}
\usepackage{arydshln}
\usepackage{algorithm}
\usepackage{subfig}
\usepackage{pifont}
\usepackage{xspace}
\usepackage{siunitx}

\title{From Dense to Sparse: Contrastive Pruning for Better Pre-trained \\ Language Model Compression}

\author{
  Runxin Xu$^{1}$\footnotemark[1], 
  Fuli Luo$^{2}$\thanks{\, Equal Contribution. Joint work between Alibaba and Peking University.}, 
  Chengyu Wang$^{2}$, Baobao Chang$^{1}$\footnotemark[2], \\
  Jun Huang$^{2}$, Songfang Huang$^{2}$\thanks{\, Corresponding authors.}, Fei Huang$^{2}$
}
\affiliations {
    $^{1}$Key Laboratory of Computational Linguistics, Peking University, MOE, China \\
    $^{2}$Alibaba Group \\
    \texttt{runxinxu@gmail.com, chbb@pku.edu.cn} \\
    \texttt{\{lfl259702,chengyu.wcy,huangjun.hj, songfang.hsf,f.huang\}@alibaba-inc.com}
}

\begin{document}

\maketitle

\newcommand{\modelname}{CAP\xspace}
\newcommand{\modelnamenew}{\textsc{\textbf{Cap}}\xspace}
\newcommand{\modelf}{\textbf{\textsc{Cap}-f}\xspace}
\newcommand{\modelm}{\textbf{\textsc{Cap}-m}\xspace}
\newcommand{\modelsoft}{\textbf{\textsc{Cap}-soft}\xspace}

\begin{abstract}
Pre-trained Language Models (PLMs) have achieved great success in various Natural Language Processing (NLP) tasks under the pre-training and fine-tuning paradigm.
With large quantities of parameters, PLMs are computation-intensive and resource-hungry.
Hence, model pruning has been introduced to compress large-scale PLMs.
However, most prior approaches only consider \textit{task-specific} knowledge towards downstream tasks, but ignore the essential \textit{task-agnostic} knowledge during pruning, which may cause catastrophic forgetting problem and lead to poor generalization ability.
To maintain both task-agnostic and task-specific knowledge in our pruned model, we propose 
\underline{\textsc{\textbf{C}}}ontr\underline{\textsc{\textbf{a}}}stive \underline{\textsc{\textbf{p}}}runing (\modelnamenew) under the paradigm of pre-training and fine-tuning.
It is designed as a general framework, compatible with both structured and unstructured pruning.
Unified in contrastive learning, \modelnamenew enables the pruned model to learn from the pre-trained model for task-agnostic knowledge, and fine-tuned model for task-specific knowledge.
Besides, to better retain the performance of the pruned model, the snapshots (i.e., the intermediate models at each pruning iteration) also serve as effective supervisions for pruning.
Our extensive experiments show that adopting \modelnamenew  consistently yields significant improvements, especially in extremely high sparsity scenarios.
With only $3\%$ model parameters reserved (i.e., $97\%$ sparsity), \modelnamenew successfully achieves $99.2\%$ and $96.3\%$ of the original BERT performance in QQP and MNLI tasks.
In addition, our probing experiments demonstrate that the model pruned by \modelnamenew tends to achieve better generalization ability.
\end{abstract}

\section{Introduction}
\begin{figure}[t]
\centering
\includegraphics[width=0.85\linewidth]{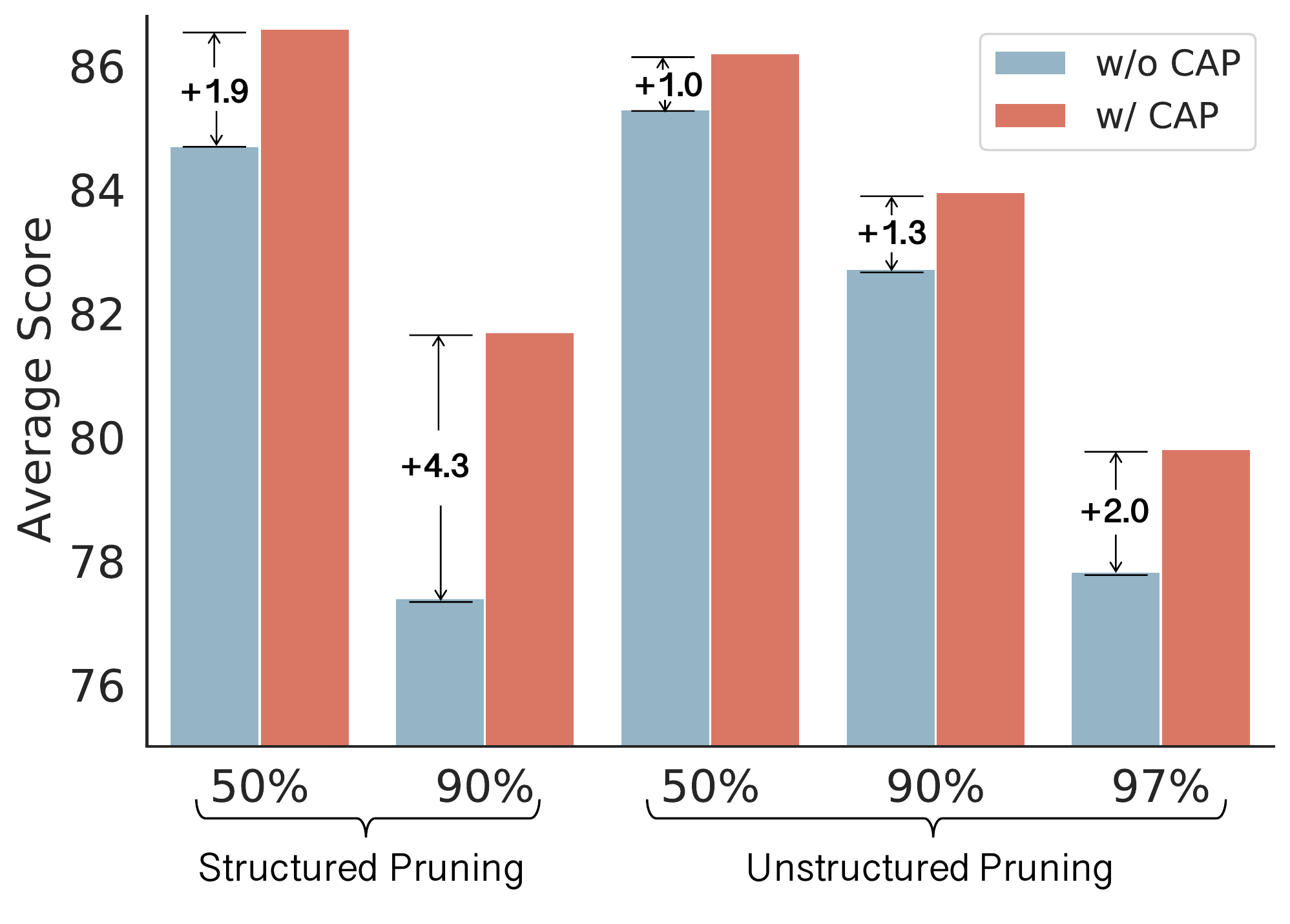}
\caption{
Comparison between BERT pruning with and without \modelnamenew.
We report the average score across MNLI, QQP, and SQuAD tasks with different model sparsity ($50\%$, $90\%$, and $97\%$).
\modelnamenew consistently yield improvements for different pruning criterions, with larger gains in higher sparsity $(1.0\rightarrow1.3\rightarrow2.0)$.
Please refer to Table~\ref{table:main} for details.
}

\label{fig:intro}
\end{figure}
Pre-trained Language Models (PLMs), such as BERT~\citep{bert}, have achieved great success in a variety of Natural Language Processing (NLP) tasks.
PLMs are pre-trained in a self-supervised way, and then adapted to the downstream tasks through fine-tuning.
Despite the success, PLMs are usually resource-hungry with a large number of parameters, ranging from millions (e.g., BERT) to billions (e.g., GPT-3), which leads to high memory consumption and computational overhead in practice.

In fact, recent studies have observed that PLMs are over-parameterized with many redundant weights~\citep{lth, ticket}. 
Motivated by this, one major line of works to compress large-scale PLMs and speed up the inference is model pruning, which focuses on identifying and removing those unimportant parameters.
However, when 
adapting the pre-trained models to downstream tasks, most studies simply adopt the vanilla pruning methods, but do not make full use of the paradigm of pre-training and fine-tuning. 
Specifically, most works only pay attention to the task-specific knowledge towards the downstream task during pruning, but ignore whether the task-agnostic knowledge of the origin PLM is well maintained in the pruned model.
Losing such task-agnostic knowledge can cause severe catastrophic forgetting problem~\citep{mixout, recadam}, which further damages the generalization ability of the pruned model.
Moreover, when facing extremely high sparsity scenarios (e.g., 97\% sparsity with only 3\% parameters reserved), the performance of the pruned model decreases sharply compared with the original dense model.

In this paper, we propose 
\underline{\textsc{\textbf{C}}}ontr\underline{\textsc{\textbf{a}}}stive \underline{\textsc{\textbf{p}}}runing (\modelnamenew), a general pruning framework under the pre-training and fine-tuning paradigm.
The core of \modelnamenew is to encourage the pruned model to learn from multiple perspectives to reserve different types of knowledge, even in extremely high sparsity scenarios.
We adopt contrastive learning~\citep{moco, simclr} to achieve the above objective with three modules: \emph{PrC}, \emph{SnC}, and \emph{FiC}. These modules contrast sentence representations derived from the pruned model with those from other models, so that the pruned model is able to learn from others and reserve corresponding representation ability. 
Specifically, \emph{PrC} and \emph{FiC} strive to pull the representation from pruned model towards that from the origin pre-trained model and fine-tuned model, to learn the task-agnostic and task-specific knowledge, respectively.
As a bridging mechanism, \emph{SnC} further strives to pull the representation from pruned model towards that from the intermediate models during pruning (called snapshots), to acquire historical and  diversified knowledge, so that the highly sparse model can still maintain comparable performance.

Our \modelnamenew framework has the following advantages:
1) \modelnamenew maintains both task-agnostic and task-specific knowledge in the pruned model, which helps alleviate catastrophic forgetting problem and maintain model performance during pruning, especially in extremely high sparsity cases;
2) \modelnamenew is based on contrastive learning that is proven to be a powerful representation learning technique;
3) \modelnamenew is a framework rather than a specific pruning method. Hence, it is orthogonal to various pruning criteria, including both structured and unstructured pruning, and can be flexibly integrated with them to offer improvements.

\modelnamenew is conceptually general and empirically powerful.
As shown in Figure~\ref{fig:intro}, our experiments show that by equipping different pruning criteria with \modelnamenew, the average scores across several tasks are consistently improved by up to $4.3$ point, achieving the state-of-the-art performance among different pruning mechanisms.
The improvement even grows larger in higher sparsity.
Our experiments also demonstrate that \modelnamenew succeeds to achieve $99.2\%$ and $96.3\%$ of the original BERT performance, with only $3\%$ model parameters in QQP and MNLI tasks.
Through task transferring probing experiments, we also find that the generalization ability of the pruned model is significantly enhanced with \modelnamenew.

\section{Background}
\subsection{Model Compression}
Pre-trained Language Models (PLMs) have achieved remarkable success in NLP community, but the demanding memory and latency also greatly increase.
Different compression methods, such as model pruning~\citep{magnitude, taylor}, knowledge distillation~\cite{tinybert, minilm}, quantization~\citep{qbert}, and matrix decomposition~\citep{albert}, have been proposed.

In this paper, we mainly focus on model pruning, which identifies and removes unimportant weights of the model. It can be divided into two categories, that is, unstructured pruning that prunes individual weights, and structured pruning that prunes structured blocks of weights.

For unstructured pruning, magnitude-based methods prunes weights according to their absolute values~\citep{magnitude, xu-etal-2021-rethinking}, while movement-based methods consider the change of weights during fine-tuning~\citep{movement}.
In addition, \citet{l0} use a hard-concrete distribution to exert L$_0$-norm regularization, and \citet{l1} introduce reweighted L$_1$-norm regularization instead.

For structured pruning, some studies use the first-order Taylor expansion to calculate the importance scores of different heads and feed-forward networks based on the variation in the loss if we remove them~\citep{taylor, sixteen, ticket, superticket}.
\citet{snip} prune modules whose  outputs are very small.
Although the above structured pruning methods are matrix-wise, there are also some studies focusing on layer-wise~\citep{layerdrop, toplayer}, and row/column-wise~\citep{schubert, block}.

Different pruning methods can be applied in a one-shot (prune for once) way, or iteratively (prune step by step) that we use in this paper.
However, most of the prior methods only consider task-specific knowledge of downstream tasks, but neglect to reserve task-agnostic knowledge in the pruned model, which leads to catastrophic forgetting problem.

\subsection{Contrastive Learning}

Contrastive learning serves as an effective mechanism for representation learning.
With similar instances considered as positive examples, and dissimilar instances as negative ones, contrastive learning aims at pulling positive examples close together and pushing negative examples apart, which usually uses InfoNCE loss~\citep{cpc}.
\citet{moco} and \citet{simclr} propose self-supervised contrastive learning in computer vision, with different views of the figure being positive examples, and different figures being negative examples.
It is also successfully introduced to NLP community, such as sentence representation~\citep{clear, simcse}, text summarization~\citep{simcls}, and so on.
In order to take advantage of annotated labels of the data, some studies extend the contrastive learning in a supervised way with an arbitrary number of positive examples~\citep{supervise-contrastive, supervise-contrastive-nlp}.

Formally, suppose that we have an example $x_i$ and it is encoded into a vector representation $z_i=\phi(x_i) \in \mathbb{R}^d $ by model $\phi$.
Besides, there are also $N$ examples being encoded into $\mathcal{S} = \left\{\hat{z}_j \right \}^{N}_{j=1}$, which are used to contrast with $z_i$.
Suppose there is one or multiple positive examples $\hat{z}_p \in \mathcal{S}$ and the others $\mathcal{S}\backslash \left \{ \hat{z}_p \right \}$ are negative examples towards $z_i$.
Following \citet{supervise-contrastive}, the contrastive training objective for example $x_i$ is defined as follows:
\begin{equation}
    \mathcal{L}_i = -\frac{1}{\left \| P(i) \right \| } \sum_{\hat{z}_j\in P(i)} \log \frac{e^{\mathrm{sim}(z_i, \hat{z}_j)/\tau}}{\sum_{k=1}^{N} e^{\mathrm{sim}(z_i, \hat{z}_k)/\tau}} 
\label{eq:multiple}
\end{equation}
where $P(i) \subset \mathcal{S}$ refers to the positive examples set for $z_i$, $\mathrm{sim}(z_i, z_j) = \frac{z_i^\top z_j}{\left \| z_i \right \| \left \| z_j \right \|} $ refers to the cosine similarity function, and $\tau$ denotes the temperature hyperparameter.

\section{Methodology}
\begin{figure}[t]
\centering
\includegraphics[width=0.95\linewidth]{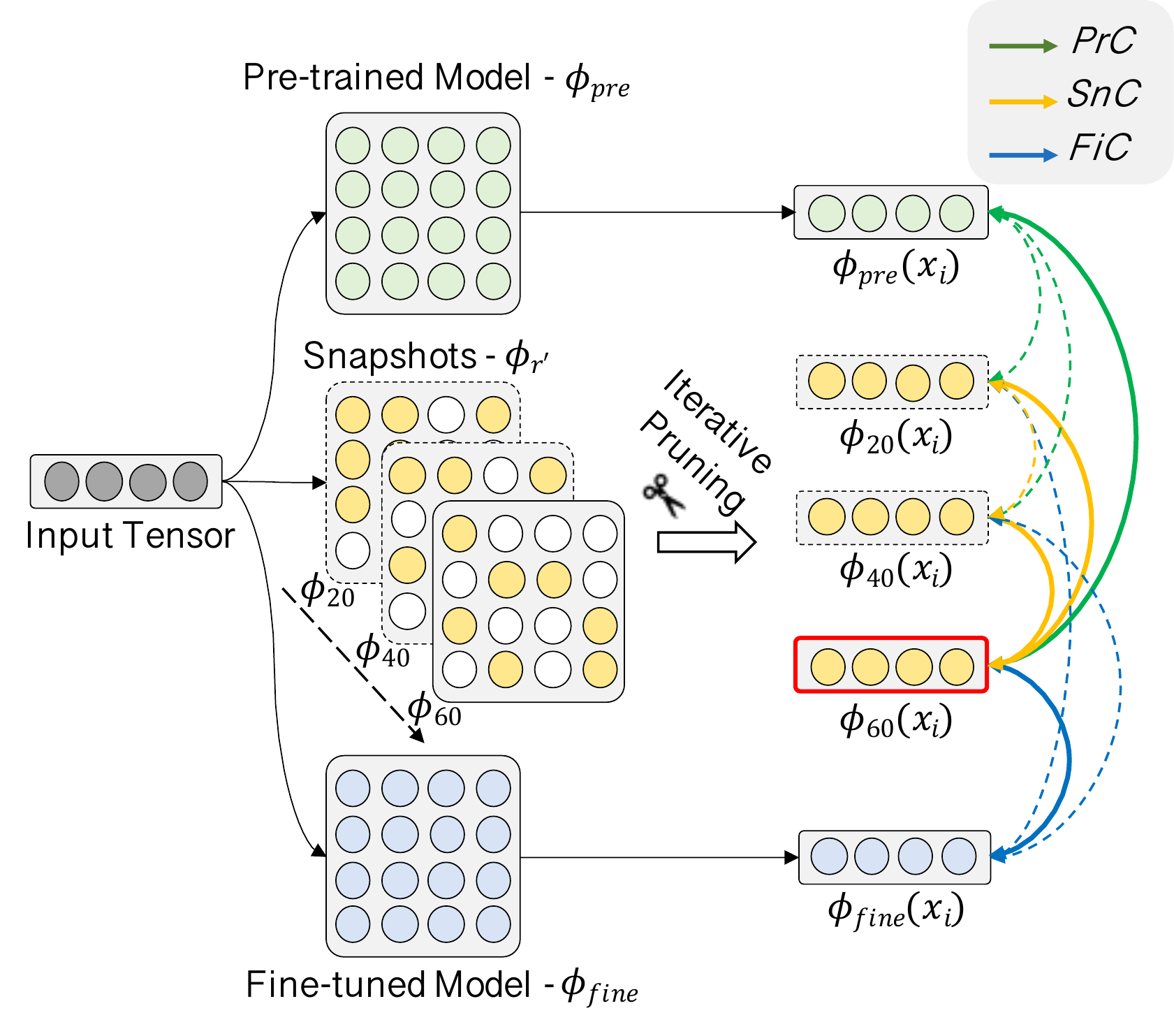}
\caption{
Overview of \modelnamenew framework,
which prunes model step by step
($\phi_{pre}\rightarrow\phi_{20}\rightarrow\phi_{40}\rightarrow\phi_{60}$), where the number denotes the sparsity ratio (\%).
Overall, \modelnamenew consists of three contrastive modules: \emph{PrC}, \emph{SnC}, and \emph{FiC}.
\emph{PrC} (green lines): contrastive learning with the pre-trained model $\phi_{pre}$ to maintain task-agnostic knowledge.
\emph{SnC} (yellow lines): contrastive learning with snapshots $\phi_{r'}$ to bridge the gap between pre-trained model and current pruned model, and gain historic and diversified knowledge.
\emph{FiC} (blue lines): contrastive learning with the fine-tuned model $\phi_{fine}$ to gain the task-specific knowledge.
The solid lines indicate the learning of the current pruned model $\phi_{60}$, while the dashed lines denote the learning of previous snapshots, $\phi_{20}$ and $\phi_{40}$.
}
\label{fig:model}
\end{figure}
\begin{table}[t]
\centering
\scalebox{0.9}{
    \begin{tabular}{lll}
    \toprule
    \bf Module & \bf Supervised & \bf Unsupervised \\
    \midrule
     \emph{PrC} & $\left \{ \phi_{pre}(x_j) \mid y_j = y_i \right \}$ & $\left \{ \phi_{pre}(x_i) \right \}$   \\
    \emph{SnC} & $\left \{ \phi_{r'}(x_j) \mid y_j = y_i, r'<r \right \}$ & $\left \{ \phi_{r'}(x_i) \mid r'<r \right \}$ \\
    \emph{FiC} & $\left \{ \phi_{fine}(x_j) \mid y_j = y_i \right \}$ & $\left \{  \phi_{fine}(x_i) \right \}$ \\
    \bottomrule
    \end{tabular}
}
\caption{
Positive examples $P(i)$ in Eq.~\ref{eq:multiple} for $\phi_{r}(x_i)$ in three contrastive modules of \modelnamenew.
$\phi_{r}(x_i)$ refers to representation of $x_i$ encoded by model $\phi_{r}$ with $r\%$ sparsity ratio, and $y_i$ denotes the annotated label for $x_i$.
}
\label{table:positive}
\end{table}

In this paper, we propose a general pruning framework, \underline{\textsc{\textbf{C}}}ontr\underline{\textsc{\textbf{a}}}stive \underline{\textsc{\textbf{p}}}runing (\modelnamenew), which prunes model via supervisions from pre-trained and fine-tuned models, and snapshots during pruning to gain different types of knowledge. 
Following iterative pruning, we compress the pre-trained model $\phi_{pre}$ to expected sparsity ratio $R\%$ progressively ($\phi_{pre} \rightarrow \phi_{r1} \rightarrow \phi_{r2} \rightarrow \cdots \rightarrow \phi_{R}$), and arbitrary pruning criteria can be used at each step. 
Figure~\ref{fig:model} illustrates the overview of \modelnamenew that consists of three modules: \emph{PrC}, \emph{SnC}, and \emph{FiC}.
They are all based on contrastive learning, with different ways to construct positive examples shown in Table~\ref{table:positive}.

\subsection{\emph{PrC}: Contrastive Learning with Pre-trained Model}
\label{sec:prc}

On the transfer learning towards a specific downstream task,
the task-agnostic knowledge in the original PLM is inclined to be lost, which can cause catastrophic forgetting problem.
Hence, in this section, we introduce a \emph{PrC} module to maintain such general-purpose language knowledge based on contrastive learning (green lines in Figure~\ref{fig:model}).

Suppose that example $x_i$ is encoded into $z_i=\phi_{r}(x_i) \in \mathbb{R}^d$ by model $\phi_{r}$ with $r\%$ sparsity ratio. 
The high-level idea is that we can contrast $z_i$ with $\left \{ \hat{z}_j=\phi_{pre}(x_j) \right \}_{j=1}^N$ encoded by pre-trained model $\phi_{pre}$ and enforce the model to correctly identify those semantically similar (positive) examples.
In this way, the current pruned model $\phi_{r}$ is able to mimic the representation modeling ability
of the pre-trained model, and therefore maintain task-agnostic knowledge.

Specifically, we adopt contrastive learning in both unsupervised and supervised settings.
For unsupervised \emph{PrC}, $\phi_{pre}(x_i)$ is considered as a positive example for $\phi_{r}(x_i)$, and $\left \{ \phi_{pre}(x_j) \right \}_{j \neq i}$ are negative examples.
The loss $\mathcal{L}_{\mathrm{unsup}}^{\mathrm{\emph{PrC}}}$ is then calculated following Eq.~\ref{eq:multiple}.
For supervised \emph{PrC}, we further utilize the sentence-level annotations of the data.
For example, the sentences are labeled as \emph{entailment}, \emph{neutral}, or \emph{contradiction} in the MNLI task.
Intuitively, we treat those having the same labels with $x_i$ as positive examples since they share similar semantic features, and the others as negative ones.
Formally, we define the positive examples set as $\left \{ \phi_{pre}(x_j) \mid y_j = y_i \right \}$, where $y_i$ denotes the label of $x_i$.
Then the supervised loss $\mathcal{L}_{\mathrm{sup}}^{\mathrm{\emph{PrC}}}$ is calculated as Eq.~\ref{eq:multiple}.
Therefore, the final training objective for \emph{PrC} is $\mathcal{L}^{\mathrm{\emph{PrC}}} = \mathcal{L}_{\mathrm{unsup}}^{\mathrm{\emph{PrC}}} + \mathcal{L}_{\mathrm{sup}}^{\mathrm{\emph{PrC}}}$.

\subsection{\emph{SnC}: Contrastive Learning with Snapshots}
\label{sec:snc}

Pruning can be applied in a one-shot or iterative way.
\emph{One-shot pruning} drops out weights and retrains the model for once.
In contrast, \emph{iterative pruning} removes weights step by step, until reaching the expected sparsity, and the intermediate models at each pruning iterations are called \textbf{snapshots}.
In this paper, we adopt iterative pruning since it better suits high sparsity regimes.
However, different from prior studies that simply ignore these snapshots, we propose \emph{SnC} to enable the current pruned model to learn from these snapshots based on contrastive learning (yellow lines in Figure~\ref{fig:model}).

In detail, suppose that we prune the model to $r\%$ sparsity ratio progressively ($\phi_{pre} \rightarrow \phi_{r1} \rightarrow \phi_{r2} \rightarrow \cdots \rightarrow \phi_{r}$), and $\left \{ \phi_{r'} \right \}_{r' < r} = \left \{ \phi_{r1}, \phi_{r2}, \dots \right \}$ are  snapshots.
Intuitively, these snapshots can bridge the gap between the sparse model ($\phi_{r}$) and dense models ($\phi_{pre}$, $\phi_{fine}$), and provide diversified supervisions with different sparse structures.
Under unsupervised settings, for example $x_i$ encoded into $\phi_{r}(x_i) \in \mathbb{R}^d$ by the current pruned model $\phi_{r}$, we treat the representations encoded from the same example but by the snapshots $\left \{ \phi_{r'}(x_i) \mid r' < r \right \}$ as positive examples, and $\left \{ \phi_{r'}(x_j) \mid j \neq i, r' < r \right \}$ as negative ones.
Under supervised settings, we utilize the annotation labels to consider instances with the same labels as positive examples.
We calculate the loss for \emph{SnC} following Eq.~\ref{eq:multiple}, $\mathcal{L}^{\mathrm{\emph{SnC}}} = \mathcal{L}_{\mathrm{unsup}}^{\mathrm{\emph{SnC}}} + \mathcal{L}_{\mathrm{sup}}^{\mathrm{\emph{SnC}}}$.
Table~\ref{table:ablation-three} show that snapshots serve as effective guidance during pruning, with $0.58\sim2.47$ average gains on MNLI, QQP, and SQuAD, especially in high sparsity regimes.

\subsection{\emph{FiC}: Contrastive Learning with Fine-tuned Model}
\label{sec:fic}

To better adapt to the downstream task, the pruned model $\phi_{r}$ can also learn from the fine-tuned model $\phi_{fine}$ that contains rich task-specific knowledge.
To this end, we propose a \emph{FiC} module, which conducts contrastive learning between the current pruned model $\phi_{r}$ and the fine-tuned model $\phi_{fine}$.
It is almost identical to the \emph{PrC} module, except that the target model $\phi_{pre}$ is replaced with the fine-tuned model $\phi_{fine}$ (blue lines in Figure~\ref{fig:model}).
Accordingly, the training loss is calculated as $\mathcal{L}^{\mathrm{\emph{FiC}}} = \mathcal{L}_{\mathrm{unsup}}^{\mathrm{\emph{FiC}}} + \mathcal{L}_{\mathrm{sup}}^{\mathrm{\emph{FiC}}}$ based on Eq.~\ref{eq:multiple}.
In addition to the contrastive supervision signal in representation space, we can also introduce distilling supervision in label space through knowledge distillation mechanism.

\subsection{Pruning with \modelnamenew Framework}

Putting \emph{PrC}, \emph{SnC}, and \emph{FiC} together, we can reach our proposed \modelnamenew framework.
Note that we can flexibly integrate with different pruning criteria in \modelnamenew.
In this paper, we try out both structured and unstructured pruning criteria.

For structured pruning, a widely used structured pruning criterion is to derive the importance score of an element based on the variation towards the loss $\mathcal{L}$ if we remove it, using the first-order Taylor expansion~\citep{taylor} of the loss.
We denote this method as \emph{First-order pruning} and absorb it into \modelnamenew, which we call \modelf.
\begin{equation}
I_w = \left | \mathcal{L}_w - \mathcal{L}_{w=0} \right |  \approx  \left | \frac{\partial \mathcal{L}}{\partial w} w \right | 
\label{eq:oneorder}
\end{equation}

For unstructured pruning, we apply the movement-based pruning methods~\citep{movement}, which calculates importance score for parameter $w$ as follows:
\begin{equation}
    I_w = -\sum_{t}\frac{\partial \mathcal{L}^{(t)}}{\partial w} w^{(t)}
\label{eq:movement}
\end{equation}
where $t$ is the training step.
Based on it, \citet{movement} reserve parameters using Top-K selection strategy or a pre-defined threshold, called \emph{Movement pruning} or \emph{Soft-movement pruning}, respectively.
We absorb these methods into \modelnamenew, and denote them as \modelm and \modelsoft.

Finally, we prune and train the model using the final objective $\mathcal{L} = \lambda_1\mathcal{L}^{CE} + \lambda_2\mathcal{L}^{PrC} + \lambda_3\mathcal{L}^{SnC} + \lambda_4\mathcal{L}^{FiC}$, where $\mathcal{L}^{CE}$ is the cross-entropy loss towards the downstream task.

\section{Extra Memory Overhead}
In our proposed \modelnamenew framework, the pruned model learns from the pre-trained, snapshots, and fine-tuned models.
However, there is unnecessary to load all of these models in GPU, which can lead to large GPU memory overhead.
In fact, because only the sentence representations of examples are required for Eq.~\ref{eq:multiple} in contrastive learning, and they also do not back-propagate the gradients, we can simply pre-encode the examples and store them in CPU.
When a normal input batch arrives, we fetch $N$ pre-encoded examples for contrastive learning.
In our paper, we use $N=4096$ by default, and the dimensions of the sentence representation for BERT$_{\mathrm{base}}$ are $768$.
Therefore, the extra GPU memory overhead is $4096\times768=3.15$M in total, and only takes up $3.15$M / $110$M = $2.86\%$ of the memory consumption of BERT$_{\mathrm{base}}$, which is low enough and acceptable.

\section{Experiments}
\subsection{Datasets}

We conduct experiments on various tasks to illustrate the effectiveness of \modelnamenew, including
a) \textbf{MNLI}, the Multi-Genre Natural Language Inference Corpus~\citep{mnli}, a natural language inference task with in-domain test set (MNLI-m), and cross-domain one (MNLI-mm).
b) \textbf{QQP}, the Quora Question Pairs dataset~\citep{glue}, a pairwise semantic equivalence task.
c) \textbf{SST-2}, the Stanford Sentiment Treebank~\citep{sst2}, a sentiment classification task for an individual sentence.
d) \textbf{SQuAD v1.1}, the Stanford Question Answering Dataset~\citep{squad}, an extractive question answering task with crowdsourced question-answer pairs.
Following most prior works~\citep{snip, movement}, we report results for the dev sets.
The detailed statistics and the metrics are provided in Appendix.

\subsection{Experiment Setups}

We conduct experiments based on BERT$_{\mathrm{base}}$~\citep{bert} with $110$M parameters, and follow their settings unless noted otherwise.
Following~\citet{movement}, we prune and report the sparsity ratio based on the weights of the encoder.
For \modelf, we prune $10\%$ parameters each step and retrain the model to recover the performance until reaching the expected sparsity.
For \modelm and \modelsoft, we follow the cubic sparsity scheduling with cool-down strategy and hyperparameter settings the same as ~\citet{movement}.
The number of examples for contrastive learning is $N=4096$.
We use the final hidden state of $[CLS]$ as the sentence representation, which is shown to be slightly better than mean pooling in our exploration experiments.
We search the temperature from $\tau=\left\{0.05, 0.1, 0.2, 0.3\right \}$.
\footnote{Our code is available at \url{https://github.com/alibaba/AliceMind/tree/main/ContrastivePruning} and \url{https://github.com/PKUnlp-icler/ContrastivePruning}.}

\subsection{Main Results}
\begin{table*}[t]
\centering
\scalebox{0.82}{
    \begin{tabular}{lcccccc}
    \toprule
    \bf Methods & \bf Sparsity & \bf MNLI-m/-mm & \bf QQP$_{\mathrm{ACC/F1}}$ & \bf SST-2 & \bf SQuAD$_{\mathrm{EM/F1}}$ \\
    \midrule
    BERT$_{\mathrm{base}}$ & 0\% & 84.5\hphantom{x}/\hphantom{x}84.4 & 90.9\hphantom{x}/\hphantom{x}88.0 & 92.9 & 80.7\hphantom{x}/\hphantom{x}88.4 \\
    \midrule
    \midrule
    
    \multicolumn{6}{c}{\emph{Knowledge Distillation}} \\
    \midrule
    \midrule
    DistillBERT~\citep{distillbert} & 50\% & 82.2\hphantom{x}/\hphantom{xxx}-\hphantom{x} & 88.5\hphantom{x}/\hphantom{xxx}-\hphantom{x} & 91.3 & 78.1\hphantom{x}/\hphantom{x}86.2 \\
    BERT-PKD~\citep{pkd} & 50\% & \hphantom{xx}-\hphantom{xx}/\hphantom{x}81.0 & 88.9\hphantom{x}/\hphantom{xxx}-\hphantom{x} & 91.5 & 77.1\hphantom{x}/\hphantom{x}85.3 \\
    TinyBERT~\citep{tinybert} & 50\% & 83.5\hphantom{x}/\hphantom{xxx}-\hphantom{x} & 90.6\hphantom{x}/\hphantom{xxx}-\hphantom{x} & 91.6 & 79.7\hphantom{x}/\hphantom{x}87.5 \\
    MiniLM~\citep{minilm} & 50\% & 84.0\hphantom{x}/\hphantom{xxx}-\hphantom{x} & 91.0\hphantom{x}/\hphantom{xxx}-\hphantom{x} & 92.0 & \hphantom{xx}-\hphantom{xx}/\hphantom{xxx}-\hphantom{x} \\
    \midrule
    TinyBERT~\citep{tinybert} & 66.7\% & 80.5\hphantom{x}/\hphantom{x}81.0 & \hphantom{xx}-\hphantom{xx}/\hphantom{xxx}-\hphantom{x} & - & 72.7\hphantom{x}/\hphantom{x}82.1 \\
    \midrule
    \midrule
    
    \multicolumn{6}{c}{\emph{Structured Pruning}} \\
    \midrule
    \midrule
    First-order~\citep{taylor} & 50\% & 83.2\hphantom{x}/\hphantom{x}83.6 & 90.8\hphantom{x}/\hphantom{x}87.5 & 90.6 & 77.2\hphantom{x}/\hphantom{x}86.0 \\
    Top-drop~\citep{toplayer} & 50\% & 81.1\hphantom{x}/\hphantom{xxx}-\hphantom{x} & 90.4\hphantom{x}/\hphantom{xxx}-\hphantom{x} & 90.3 & \hphantom{xx}-\hphantom{xx}/\hphantom{xxx}-\hphantom{x} \\
    SNIP~\citep{snip} & 50\% & \hphantom{xx}-\hphantom{xx}/\hphantom{x}82.8 & 88.9\hphantom{x}/\hphantom{xxx}-\hphantom{x} & 91.8 & \hphantom{xx}-\hphantom{xx}/\hphantom{xxx}-\hphantom{x} \\
    schuBERT~\citep{schubert} & 50\% & 83.8\hphantom{x}/\hphantom{xxx}-\hphantom{x} & \hphantom{xx}-\hphantom{xx}/\hphantom{xxx}-\hphantom{x} & 91.7 & 80.7\hphantom{x}/\hphantom{x}88.1 \\
    \modelf (Ours) & 50\% & \bf 84.5\hphantom{x}/\hphantom{x}85.0 & \bf 91.6\hphantom{x}/\hphantom{x}88.6 & \bf 92.7 & \bf 81.4\hphantom{x}/\hphantom{x}88.7 \\
    \midrule
    SNIP~\citep{snip} & 75\% & \hphantom{xx}-\hphantom{xx}/\hphantom{x}78.3 & 87.8\hphantom{x}/\hphantom{xxx}-\hphantom{x} & 88.4 & \hphantom{xx}-\hphantom{xx}/\hphantom{xxx}-\hphantom{x} \\
    First-order~\citep{taylor} & 90\% & 79.1\hphantom{x}/\hphantom{x}79.5 & 88.7\hphantom{x}/\hphantom{x}84.9 & 86.9 & 59.8\hphantom{x}/\hphantom{x}72.3 \\
    \modelf (Ours) & 90\% & \bf 81.0\hphantom{x}/\hphantom{x}81.2 & \bf 90.2\hphantom{x}/\hphantom{x}86.9 & \bf 89.7 & \bf 70.2\hphantom{x}/\hphantom{x}80.6 \\
    \midrule 
    \midrule
    
    \multicolumn{6}{c}{\emph{Unstructured Pruning}} \\
    \midrule
    \midrule
    Movement~\citep{movement} & 50\% & 82.5\hphantom{x}/\hphantom{x}82.9 & 91.0\hphantom{x}/\hphantom{x}87.8 & - & 79.8\hphantom{x}/\hphantom{x}87.6 \\
    \modelm (Ours) & 50\% & \bf 83.8\hphantom{x}/\hphantom{x}84.2 & \bf 91.6\hphantom{x}/\hphantom{x}88.6 & - & \bf 80.9\hphantom{x}/\hphantom{x}88.2 \\
    \midrule
    Magnitude~\citep{magnitude} & 90\% & 78.3\hphantom{x}/\hphantom{x}79.3 & 79.8\hphantom{x}/\hphantom{x}65.0 & - & 70.2\hphantom{x}/\hphantom{x}80.1 \\
    L$_0$-regularization~\citep{l0} & 90\% & 78.7\hphantom{x}/\hphantom{x}79.7 & 88.1\hphantom{x}/\hphantom{x}82.8 & - & 72.4\hphantom{x}/\hphantom{x}81.9 \\
    Movement~\citep{movement} & 90\% & 80.1\hphantom{x}/\hphantom{x}80.4 & 89.7\hphantom{x}/\hphantom{x}86.2 & - & 75.6\hphantom{x}/\hphantom{x}84.3 \\
    Soft-Movement~\citep{movement} & 90\% & 81.2\hphantom{x}/\hphantom{x}81.8 & 90.2\hphantom{x}/\hphantom{x}86.8 & - & 76.6\hphantom{x}/\hphantom{x}84.9 \\
    \modelm (Ours) & 90\% & 81.1\hphantom{x}/\hphantom{x}81.8 & 91.6\hphantom{x}/\hphantom{x}87.7 & - & 76.5\hphantom{x}/\hphantom{x}85.1 \\
    \modelsoft (Ours) & 90\% & \bf 82.0\hphantom{x}/\hphantom{x}82.9 & \bf 90.7\hphantom{x}/\hphantom{x}87.4 & - & \bf 77.1\hphantom{x}/\hphantom{x}85.6 \\
    \midrule
    Movement~\citep{movement} & 97\% & 76.5\hphantom{x}/\hphantom{x}77.4 & 86.1\hphantom{x}/\hphantom{x}81.5 & - & 67.5\hphantom{x}/\hphantom{x}78.0 \\
    Soft-Movement~\citep{movement} & 97\% & 79.5\hphantom{x}/\hphantom{x}80.1 & 89.1\hphantom{x}/\hphantom{x}85.5 & - & 72.7\hphantom{x}/\hphantom{x}82.3 \\
    \modelm (Ours) & 97\% & 77.5\hphantom{x}/\hphantom{x}78.4 & 88.8\hphantom{x}/\hphantom{x}85.0 & - & 69.5\hphantom{x}/\hphantom{x}79.7 \\
    \modelsoft (Ours) & 97\% & \bf 80.1\hphantom{x}/\hphantom{x}81.3 & \bf 90.2\hphantom{x}/\hphantom{x}86.7 & - & \bf 73.8\hphantom{x}/\hphantom{x}83.0 \\
    \bottomrule
    \end{tabular}
}
\caption{
Comparison between \modelnamenew with other model compression methods without data augmentation.
\modelnamenew consistently achieve the best performance under the same sparsity ratio across different tasks.
With only $3\%$ of the encoder's parameter (i.e., 97\% sparsity), \modelsoft still reaches $99.2\%$ and $96.3\%$ of the original BERT performance in QQP and MNLI task, respectively.
}
\label{table:main}
\end{table*}

In this section, we compare \modelnamenew with the following model compression methods:
1) \emph{knowledge distillation}: DistillBERT~\citep{distillbert}, BERT-PKD~\citep{pkd}, TinyBERT~\citep{tinybert}, and MiniLM~\citep{minilm}.
Note that we report the results of TinyBERT without data augmentation mechanism to ensure fairness.
2) \emph{structured pruning}:
the most standard First-order pruning~\citep{taylor} that  \modelf is based, Top-drop~\citep{toplayer}, SNIP~\citep{snip}, and schuBERT~\citep{schubert}.
3) \emph{unstructured pruning}: Magnitude pruning~\cite{magnitude}, L$_0$-regularization~\citep{l0}, and the state-of-the-art Movement pruning and Soft-movement pruning~\citep{movement} that our \modelm and \modelsoft are based on.
Please refer to the Appendix for more details about the baselines.
Table~\ref{table:main} illustrates the main results, from which we have some important observations.

(1) \emph{\modelnamenew removes a large proportion of BERT parameters while still maintaining comparable performance}.
With $50\%$ sparsity ratio, \modelf achieves an equal score in MNLI-m compared with origin BERT, and even improves by $0.3\sim0.7$ score for MNLI-mm, QQP, and SQuAD tasks.
More importantly, with only $3\%$ of the encoder's parameters (i.e., $97\%$ sparsity ratio), our \modelsoft successfully achieves $99.2\%$ and $96.3\%$ of the original BERT performance in QQP ($90.9\rightarrow90.2$) and MNLI-mm ($84.4\rightarrow81.3$).

(2) \emph{\modelnamenew consistently yields improvements for different pruning criteria, along with larger gains in higher sparsity}.
Compared with structured First-order pruning, \modelf improves by $1.9 (85.6\rightarrow87.5)$ and $4.1 (78.7\rightarrow82.8)$ average score over all tasks under $50\%$ and $90\%$ sparsity.
Similarly, compared with unstructured Movement pruning, as the sparsity grows by $50\%\rightarrow90\%\rightarrow97\%$, \modelm improves $1.0 (85.2\rightarrow86.2)$, $1.3 (82.7\rightarrow84.0)$, and $2.0 (77.8\rightarrow79.8)$ scores, which also increases accordingly.

(3) \emph{\modelnamenew consistently outperforms other pruning methods}.
For example, \modelf surpasses SNIP by $2.2$ and $2.7$ score in MNLI-mm and QQP tasks under $50\%$ sparsity.
Besides, with higher sparsity, the $90\%$-sparsified \modelf can even beat SNIP with $75\%$ sparsity, with $2.9$ and $2.4$ higher score in the MNLI-mm and QQP tasks.

(4) \emph{\modelnamenew also outperforms knowledge distillation methods}.
For example, compared with TinyBERT under $66.7\%$ sparsity ratio, \modelf that is under $90\%$ sparsity ratio can still surpass it by $0.5$ accuracy in the MNLI-m task.

The above observations support our claim that \modelnamenew helps the pruned model to maintain both task-agnostic and task-specific knowledge and hence benefits the pruning, especially under extremely high sparsity scenarios.

\subsection{Generalization Ability}
\begin{figure}[t]
\centering
\includegraphics[width=0.95\linewidth]{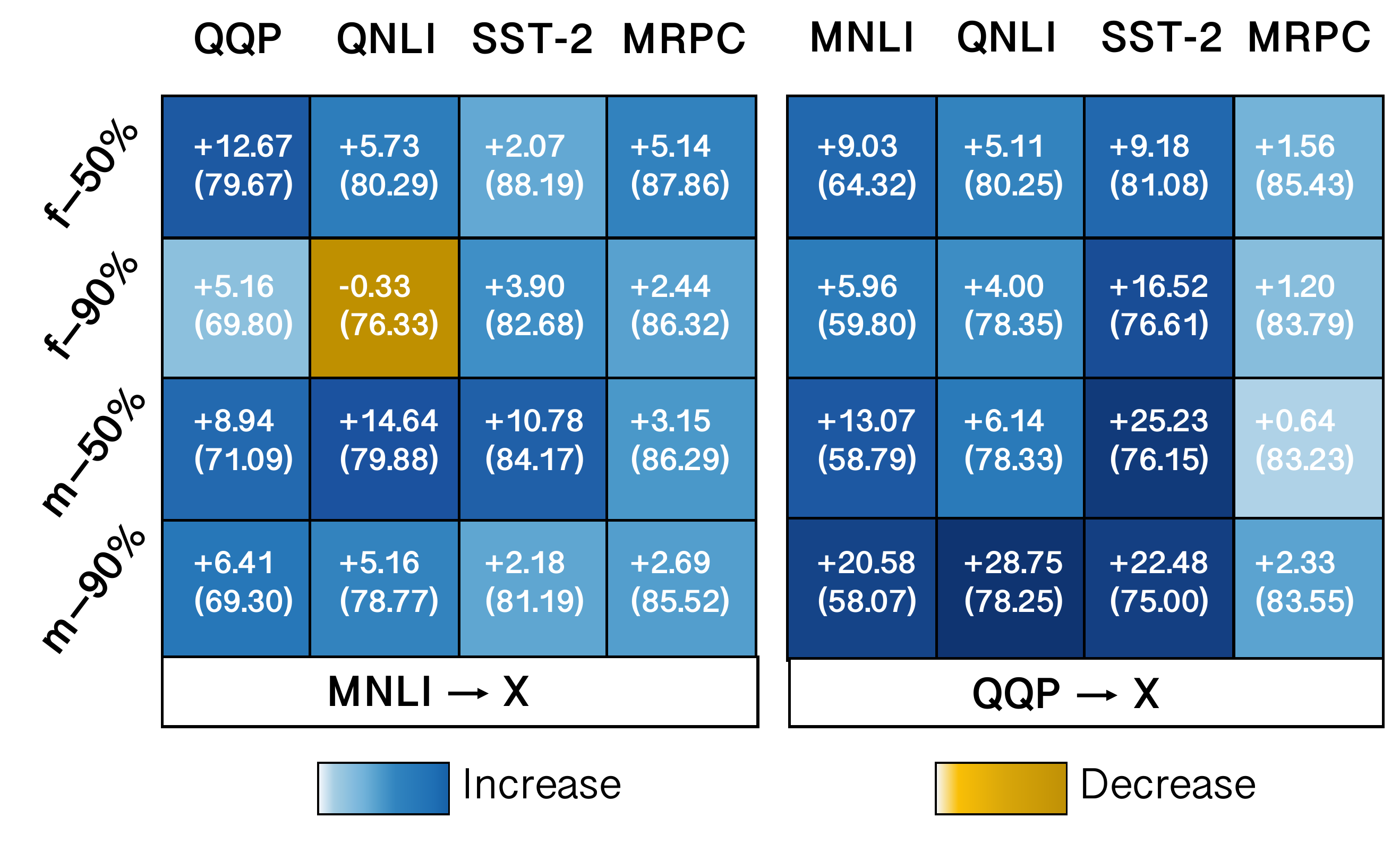}
\caption{
Generalization ability probing.
We transfer the pruned model trained on MNLI (left) and QQP (right) to target tasks under $50\%$ and $90\%$ sparsity ratio.
We report the improvement brought by \modelnamenew compared with its basic pruning method, and the absolute performance of \modelnamenew (in bracket).
\modelnamenew yields improvement in most cases, suggesting better generalization ability of the model pruned by \modelnamenew.
}
\label{fig:transfer}
\end{figure}

Different from prior methods, \modelnamenew can maintain task-agnostic knowledge in the pruned model, and therefore strengthen its generalization ability.
To justify our claim, we follow the task transfer probing experiments from ~\citet{rxf}.
In detail, we freeze the representations derived from the pruned model trained on MNLI or QQP, and then only train a linear classifier on top of the model for another task.
Besides MNLI, QQP, and SST-2 tasks we have used, we also include QNLI~\citep{glue} and  MRPC~\citep{mrpc} as our target tasks\footnote{We report F1 for QQP/MRPC and accuracy for other tasks.}.

The results under $50\%$ and $90\%$ sparsity are shown in Figure~\ref{fig:transfer}, where the first two rows correspond to the improvement of \modelf over First-order pruning, and the last two rows correspond to the improvement of \modelm over Movement pruning.
Improvements are shown at each cell, with the performance score of \modelnamenew in the bracket.
As is shown, \modelnamenew yields improvements in an overwhelming majority of cases.
For example, \modelm outperforms Movement pruning by a large margin, with up to $28.75$ higher score in QNLI transferred from QQP task, under $90\%$ sparsity.
The significant improvements in task transferring experiments suggest \modelnamenew can better maintain the task-agnostic knowledge and strengthen the generalization ability of the pruned model.

\section{Discussions}
\subsection{Understanding Different Contrastive Modules}
\label{sec:ablation-three}

\begin{table}[t]
\centering
\scalebox{0.8}{
\begin{tabular}{lccccc}
\toprule
\bf Methods  & \bf Sparsity          & \bf MNLI-m & \bf QQP & \bf SQuAD &    $\bf \Delta$   \\
\midrule
\modelf   & 50\%     & 84.48  & 88.49  & 81.37    &   -    \\
- w/o \emph{PrC} &     & -0.47  & -0.50   & -0.67    & -0.55 \\
- w/o \emph{SnC} &   & -0.47  & -0.57  & -1.10     & -0.71 \\
- w/o \emph{FiC} &   & -0.42  & -0.63  & -0.58    & -0.54 \\
\midrule
\modelf    & 90\%        & 80.98  & 86.92  & 70.16    &    -   \\
- w/o \emph{PrC} &     & -1.78  & -0.87  & -5.11    & -2.59 \\
- w/o \emph{SnC} &   & -1.57  & -1.72  & -4.11    & -2.47 \\
- w/o \emph{FiC} &     & -1.53  & -1.31  & -2.64    & -1.83 \\
\midrule
\modelm    & 50\%        & 83.29  & 88.28  & 80.40     &   -    \\
- w/o \emph{PrC} &     & -0.84  &  -0.22      & -0.74     & -0.60 \\
- w/o \emph{SnC} &   & -0.74  & -0.36  & -0.66    & -0.59 \\
- w/o \emph{FiC} &     & -1.31  & -0.77  & -1.07    & -1.05 \\
\midrule
\modelm    & 90\%        & 80.53  & 87.12  & 76.20     &   -    \\
- w/o \emph{PrC} &     & -0.58  &   -0.47     & -0.68    & -0.58 \\
- w/o \emph{SnC} &   & -0.45  & -0.78  & -0.66    & -0.58 \\
- w/o \emph{FiC} &     & -1.54  & -0.73  & -0.94    & -1.07 \\
\midrule
\modelm    & 97\%        & 77.30   & 84.70   & 69.47    &    -   \\
- w/o \emph{PrC} &     & -0.27  &   -0.23     & -1.87    & -0.79 \\
- w/o \emph{SnC} &   & -0.29  & -0.52  & -1.67   & -0.83 \\
- w/o \emph{FiC} &     & -1.31  & -0.34  & -2.11    & -1.25 \\
\bottomrule
\end{tabular}
}
\caption{
Ablation study of different contrastive modules. 
$\Delta$ refers to the average score reduction across all tasks.
Removing any contrastive module would cause degradation of the pruned model, especially in highly sparse regimes.
}
\label{table:ablation-three}
\end{table}

\modelnamenew is mainly comprised of three major contrastive modules, \emph{PrC} for learning from the pre-trained model, \emph{SnC} for learning from snapshots, and \emph{FiC} for learning from the fine-tuned model.
To better explore their effects, we remove one of them from \modelnamenew at a time, and prunes model with \modelf and \modelm methods under various sparsity ratios.

We report the accuracy for MNLI-m, F1 for QQP, and Exact Match score for SQuAD, along with the average score decrease $\Delta$.
As shown in Table~\ref{table:ablation-three}, 
removing any contrastive module would cause degradation of the model.
For example, with $90\%$ sparsity ratio, removing \emph{PrC}, \emph{SnC}, or \emph{FiC} leads to $1.83\sim2.59$ average score reduction for \modelf, and $0.58\sim1.07$ for \modelm.
Besides, the performance degradation gets larger as the sparsity gets higher in general.
This is especially evident for \modelf.
For instance, using \modelf without \emph{PrC} module, when the sparsity ratio varies from $50\%$ to $90\%$, the degradation increases from $0.55$ to $2.59$, 
The results demonstrate that all  contrastive modules play important roles in \modelnamenew, and have complementary advantages with each other, especially in highly sparse regimes.

\subsection{Understanding Supervised and Unsupervised Contrastive Objectives}
\label{sec:ablation-contrastive}

\begin{table}[t]
\centering
\scalebox{0.76}{
\begin{tabular}{lcccc}
\toprule
\bf Methods         & \bf Sparsity & \bf MNLI-m/-mm & \bf QQP$_{\mathrm{ACC/F1}}$ & $\bf \Delta$   \\
\midrule
\modelf          & 50\%     & 84.48\hphantom{x}/\hphantom{x}84.97   & 91.45\hphantom{x}/\hphantom{x}88.49  &     -    \\
- w/o sup    &          & -0.64\hphantom{x}/\hphantom{x}-0.30    & -0.53\hphantom{x}\hphantom{x}/-0.65  & -0.53   \\
- w/o unsup &          & -0.47\hphantom{x}/\hphantom{x}-0.24   & -0.56\hphantom{x}/\hphantom{x}-0.69  & -0.49   \\
\midrule
\modelf          & 90\%     & 80.98\hphantom{x}/\hphantom{x}81.17   & 90.23\hphantom{x}/\hphantom{x}86.92  &     -    \\
- w/o sup    &          & -0.97\hphantom{x}/\hphantom{x}-0.69   & -0.84\hphantom{x}/\hphantom{x}-0.99  & -0.87 \\
- w/o unsup &          & -1.29\hphantom{x}/\hphantom{x}-1.03   & -0.85\hphantom{x}/\hphantom{x}-1.19  & -1.09   \\
\midrule
\modelm          & 50\%     & 83.29\hphantom{x}/\hphantom{x}83.91   & 91.34\hphantom{x}/\hphantom{x}88.28  &     -    \\
- w/o sup    &          & -0.80\hphantom{x}/\hphantom{x}-0.76   & -0.37\hphantom{x}/\hphantom{x}-0.47  & -0.60    \\
- w/o unsup &          & -0.86\hphantom{x}/\hphantom{x}-0.53   & -0.48\hphantom{x}/\hphantom{x}-0.63  & -0.63  \\
\midrule
\modelm          & 90\%     & 80.53\hphantom{x}/\hphantom{x}81.13   & 90.44\hphantom{x}/\hphantom{x}87.12  &    -     \\
- w/o sup    &          & -0.71\hphantom{x}/\hphantom{x}-0.85   & -1.36\hphantom{x}/\hphantom{x}-0.70   & -0.91  \\
- w/o unsup &          & -0.91\hphantom{x}/\hphantom{x}-1.18   & -1.62\hphantom{x}/\hphantom{x}-1.29  & -1.25   \\
\midrule
\modelm          & 97\%     & 77.30\hphantom{x}/\hphantom{x}78.21   & 88.56\hphantom{x}/\hphantom{x}84.7   &    -     \\
- w/o sup    &          & -0.36\hphantom{x}/\hphantom{x}-0.43   & -0.95\hphantom{x}/\hphantom{x}-1.52  & -0.82  \\
- w/o unsup &          & -0.39\hphantom{x}/\hphantom{x}-0.58   & -0.80\hphantom{x}/\hphantom{x}-1.02  & -0.70 \\
\bottomrule
\end{tabular}
}
\caption{
Ablation study of supervised (sup) and unsupervised (unsup) training objectives. 
$\Delta$ is the average score reduction across all scores.
Both supervised and unsupervised contrastive learning objectives are beneficial to the pruning, and combining them can lead to better performance.
}
\label{table:ablation-contrastive}
\end{table}

In \modelnamenew, the same example encoded by different models are considered as positive examples for unsupervised contrastive learning (unsup).
If the sentence-level label annotations are available, we can also conduct supervised contrastive learning (sup) by considering examples with the same labels as positive examples.
To explore their effects, we remove either of them for the ablation study.

As demonstrated in Table~\ref{table:ablation-contrastive}, without either supervised or unsupervised contrastive learning objectives, the performance of the pruned model would markedly decline.
Specifically, removing the supervised contrastive learning objective leads to $0.53\sim0.91$ average score decrease for \modelnamenew, while abandoning the unsupervised one also causes $0.49\sim1.25$ decrease.
It suggests that both supervised and unsupervised objectives are essential and necessary for \modelnamenew, and their advantages are orthogonal to each other.

\subsection{Performance under Various Sparsity Ratios}

\begin{figure}[t]
\centering
\subfigure[QQP]{
\begin{minipage}[t]{0.46\linewidth}
\centering
\includegraphics[width=\linewidth]{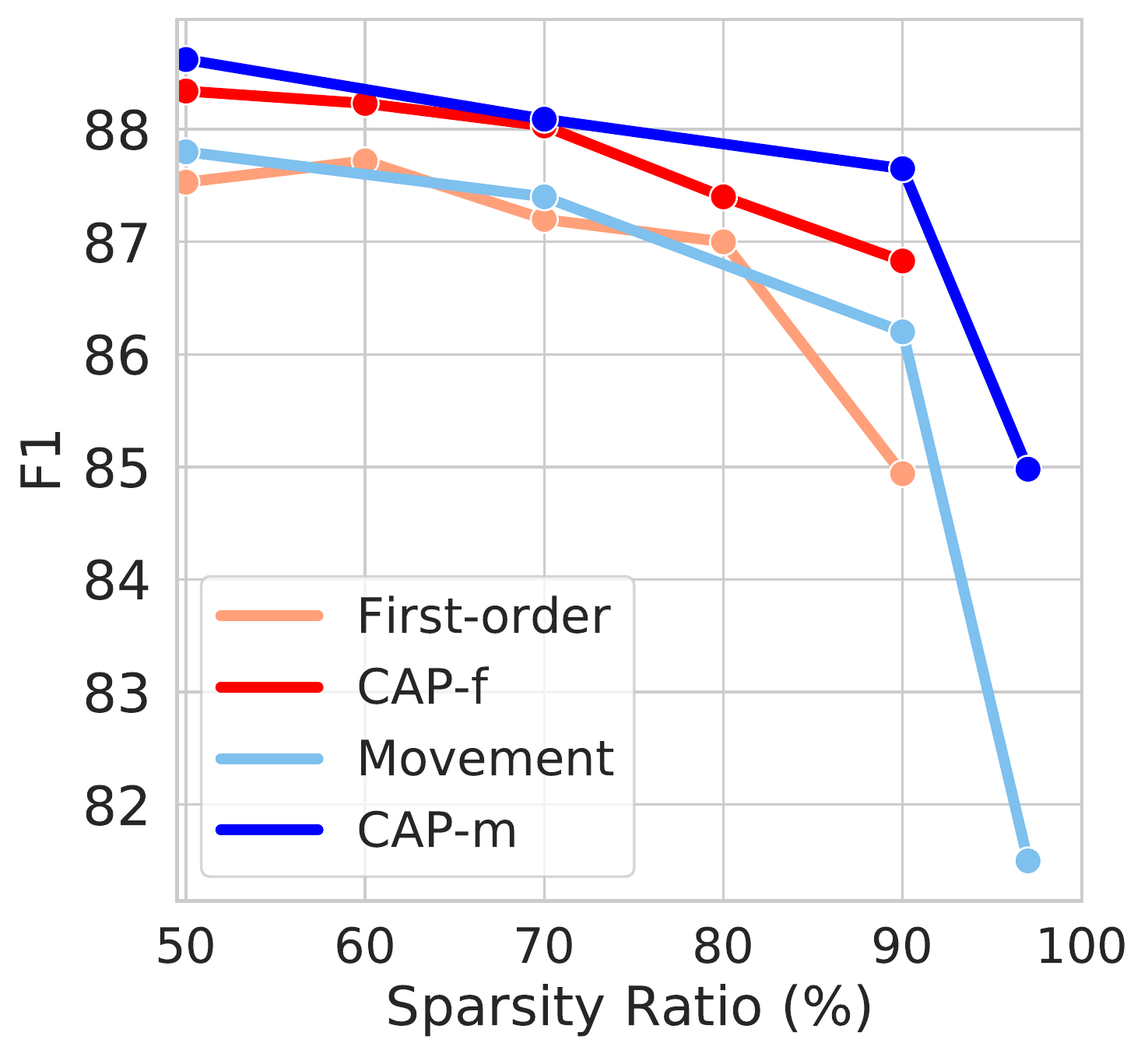}
\end{minipage}
}
\subfigure[SQuAD]{
\begin{minipage}[t]{0.46\linewidth}
\centering
\includegraphics[width=\linewidth]{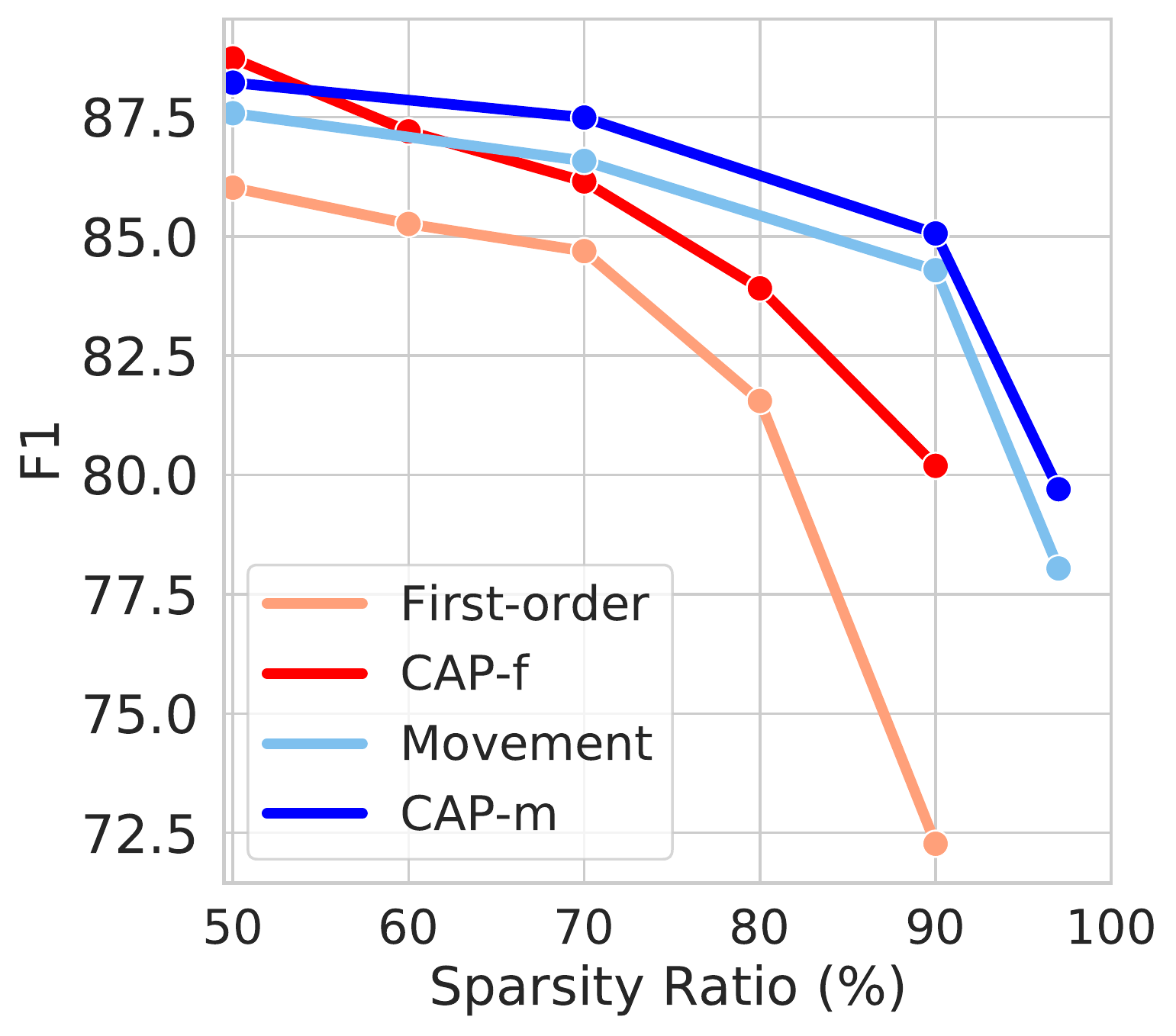}
\end{minipage}
}
\centering
\caption{
Comparison between \modelnamenew and its basic pruning methods.
\modelf consistently outperforms First-order pruning, and \modelm also surpasses Movement pruning under various sparsity ratios.
}
\label{fig:ratio}
\end{figure}

In this section, we compare \modelf and \modelm with their basic pruning methods, First-order pruning and Movement pruning, with sparsity ratios varying from $50\%$ to $97\%$.
The evolution of the performance in QQP and SQuAD task is shown in Figure~\ref{fig:ratio}.
The performance of the pruned model for all methods decreases as the sparsity ratio increases.
However, we can observe that \modelf consistently outperforms First-order pruning, and the improvement is even larger in higher sparsity situations.
A similar tendency also exists between \modelm and Movement pruning.
These results suggest that with three core contrastive modules, \modelnamenew can better maintain the model performance during the pruning.

\subsection{Exploration of Pooling Methods and Temperatures}

\begin{figure}[t]
\centering
\subfigure[\modelf]{
\begin{minipage}[t]{0.46\linewidth}
\centering
\includegraphics[width=\linewidth]{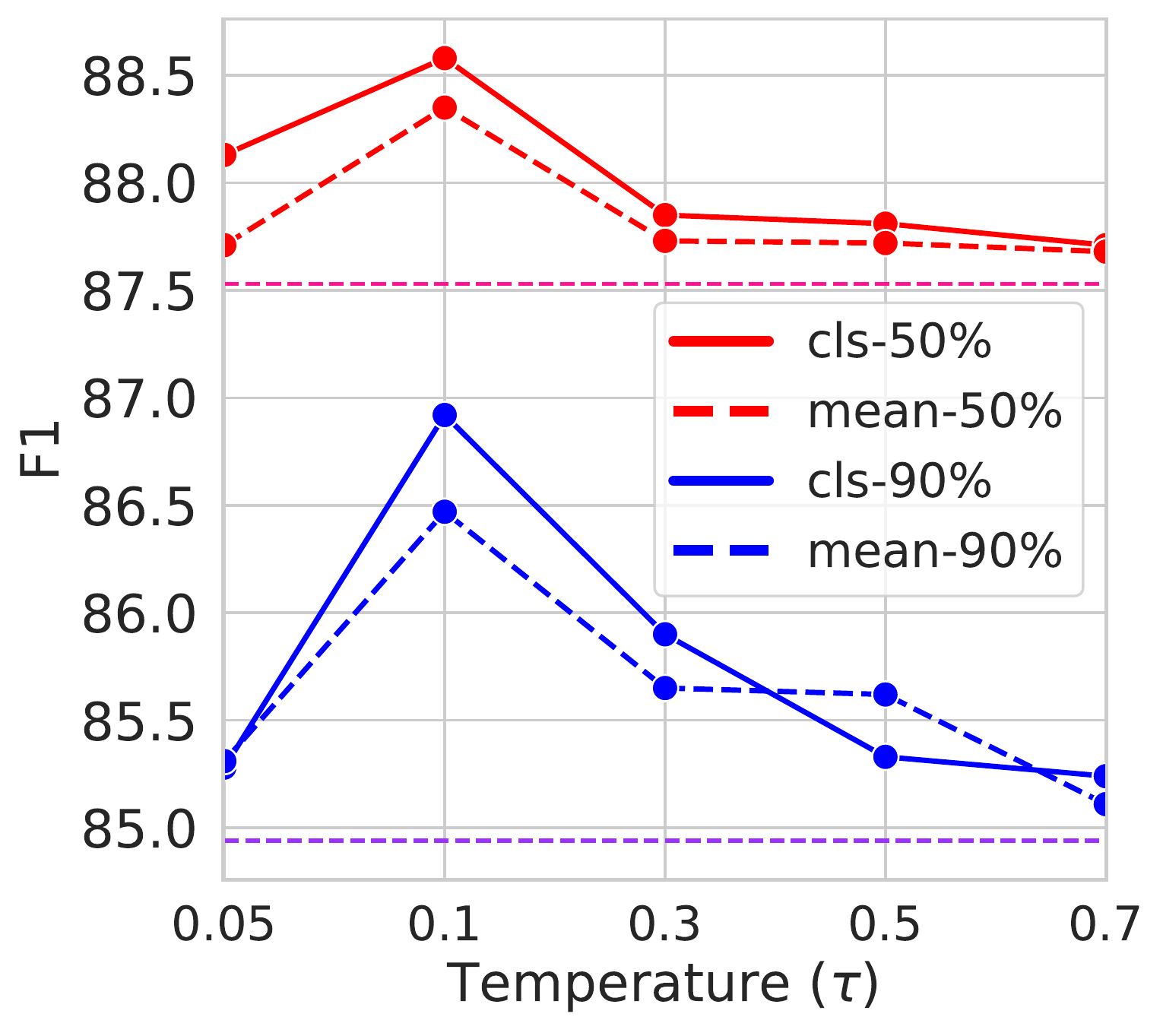}
\end{minipage}
}
\subfigure[\modelm]{
\begin{minipage}[t]{0.46\linewidth}
\centering
\includegraphics[width=\linewidth]{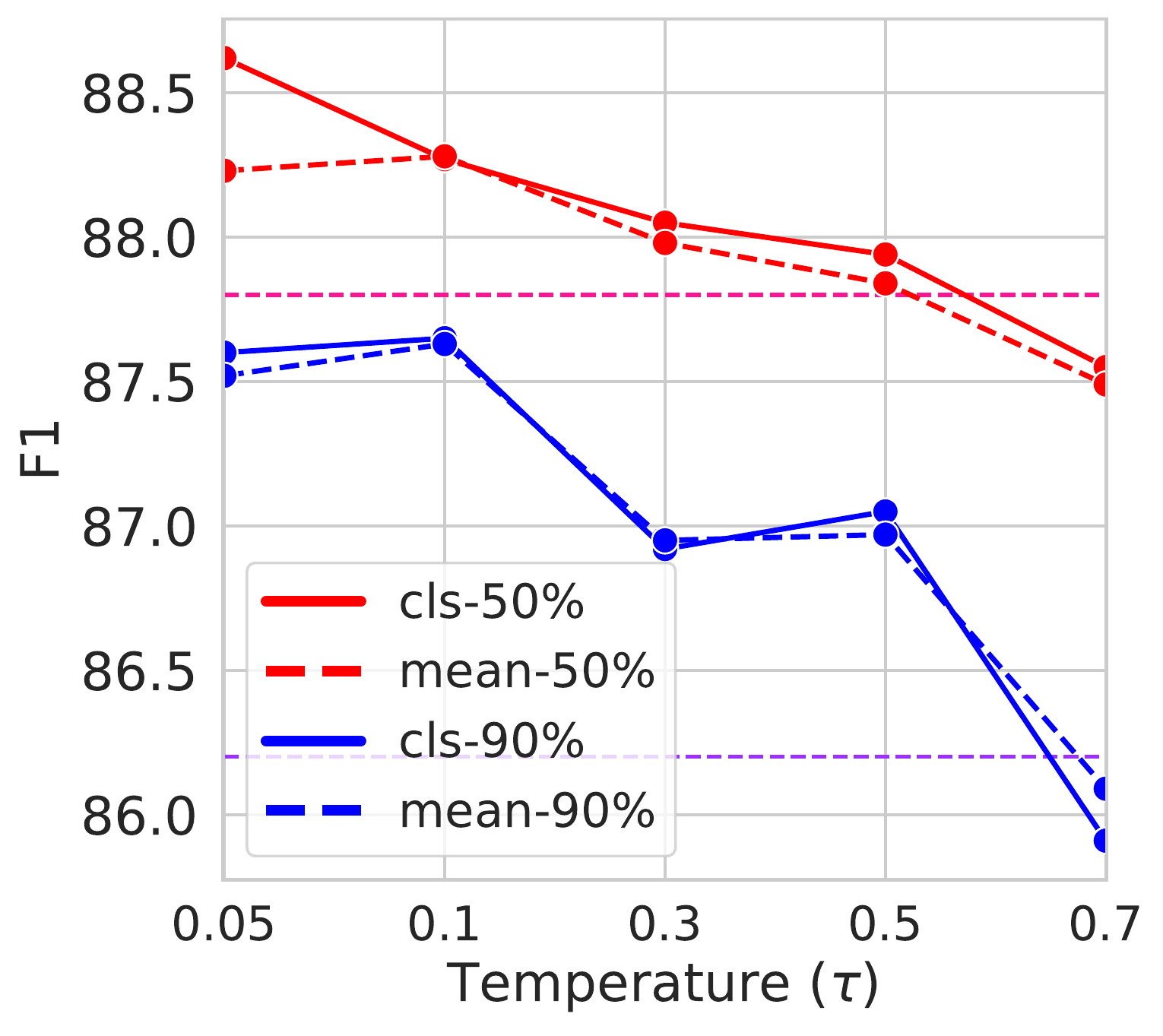}
\end{minipage}
}
\centering
\caption{
Performance of different pooling methods and temperatures.
The horizontal dashed line denotes the performance of First-order and Movement pruning under $50\%$ and $90\%$ sparsity.
In general, using $[CLS]$ as sentence representations slightly outperforms mean pooling, and temperature $\tau=0.1$ tends to achieve the best results.
}
\label{fig:temperature}
\end{figure}

The contrastive loss function in Eq.~\ref{eq:multiple} involves two important points, the sentence representation $z_i$ and the temperature $\tau$.
For sentence representation $z_i$, we explore two pooling methods of the hidden states encoded by the model, using the vector representation of the $[CLS]$ or the mean pooling of all representations of the whole sentence.
For temperature $\tau$, we also explore different values ranging from $0.05$ to $0.7$.
We conduct experiments on QQP.
As shown in Figure~\ref{fig:temperature}, using $[CLS]$ as sentence representations slightly outperforms the mean pooling method.
Besides, \modelnamenew can achieve better performance than its basic pruning method under most temperatures, and setting $\tau=0.1$ tends to achieve the best performance in most cases.

\subsection{Exploration of Learning From Fine-tuned Model}

\begin{table}[t]
\centering
\scalebox{0.8}{
    \begin{tabular}{lcccc}
    \toprule
    \multirow{2}{*}{\bf Methods} & \multicolumn{2}{c}{\bf QQP} & 
    \multicolumn{2}{c}{\bf SQuAD} \\
    \cmidrule(lr){2-3} \cmidrule(lr){4-5}
    ~ & 50\% & 90\% & 50\% & 90\% \\
    \midrule
    Movement & 87.80 & 86.20 & 87.58 & 84.29 \\
    \quad - w/o KD & 87.30 & 83.20 & 83.16 & 81.72 \\
    
    \midrule
    \modelf & 88.58 & 86.92 & 88.73 & 80.59 \\
    \quad - w/o KD & 88.59 & 86.88 & 86.52 & 77.76 \\
    
    \midrule
    \modelm & 88.62 & 87.65 & 88.22 & 85.06 \\
    \quad - w/o KD & 88.60 & 87.67 & 85.94 & 82.46 \\
    
    \bottomrule
    \end{tabular}
}
\caption{
Exploring of learning from the fine-tuned model. 
For \modelnamenew, KD brings improvements for token-level task (SQuAD), but has little effect on sentence-level task (QQP).
}
\label{table:kd}
\end{table}

To gain task-specific knowledge, we propose to perform contrastive learning on the sentence representations from the fine-tuned model (\emph{FiC}).
Another common way is to perform the knowledge distillation (KD) on the soft label that has already been shown effective in pruning~\citep{movement, dynabert}.
To explore the effect of KD, we conduct further experiments in Table~\ref{table:kd}. 
It shows that KD boosts the performance of CAP in token-level task (SQuAD) while has little effect on sentence-level task (QQP). 
The reason can be that the contrastive learning on the sentence representations is sufficient to capture the features of the sentence-level task, while the information still incurs losses on token-level tasks. 
Thus, for token-level tasks, it is better to perform \modelnamenew with KD.

\section{Conclusion}
In this paper, we propose a general pruning framework, \underline{\textsc{\textbf{C}}}ontr\underline{\textsc{\textbf{a}}}stive \underline{\textsc{\textbf{p}}}runing (\modelnamenew), under the paradigm of pre-training and fine-tuning.
Based on contrastive learning, we enhance the pruned model to maintain both task-agnostic and task-specific knowledge via pulling it towards the representations from the pre-trained model $\phi_{pre}$, and fine-tuned model $\phi_{fine}$. 
Furthermore, the snapshots during the pruning process are also fully utilized to provide historic and diversified supervisions to retain the performance of the pruned model, especially in high sparsity regimes.
\modelnamenew consistently yields significant improvements to different pruning criteria, and achieves the state-of-the-art performance among different pruning mechanisms.
Experiments also show that \modelnamenew strengthen the generalization ability of the pruned model.

\section*{Acknowledgments}

This paper is supported by the National Key Research and Development Program of China under Grant No. 2020AAA0106700, the National Science Foundation of China under Grant No.61936012 and 61876004.

\bibliography{aaai22}

\clearpage
\appendix
\subsection{Datasets and Metrics}
\label{appendix:dataset}
The detailed statistics of different datasets are shown in Table~\ref{table:dataset}.
The metrics we report for each dataset are also illustrated in Table~\ref{table:dataset}.
We obtain the data from \url{https://huggingface.co/datasets/glue}.

\begin{table}[htbp]
\centering
\begin{tabular}{lccc}
\toprule
\bf Dataset & \bf \#Train & \bf \#Dev & \bf Metrics \\
\midrule
MNLI-m & 393k & 9.8k & Accuracy \\
MNLI-mm & 393k & 9.8k & Accuracy \\
QQP & 364k & 40k & Accuracy/F1 \\
SST-2 & 67k & 872 & Accuracy \\
SQuAD & 88K & 11k & Exact Match/F1 \\
\bottomrule
\end{tabular}
\caption{Statistics and metrics of datasets used in this paper.
}
\label{table:dataset}
\end{table}

\subsection{Baseline Compression Methods}

We compare with the following compression methods.

\paragraph{Knowledge Distillation}
\begin{itemize}
    \item \textbf{DistillBERT}~\citep{distillbert}, which distills the soft label on the downstream task predicted by the teacher model towards the student model.
    \item \textbf{BERT-PKD}~\citep{pkd}, which also distills the hidden states of the teacher model towards the student models in representation space.
    \item \textbf{TinyBERT}~\citep{tinybert}, which distills the  hidden states and soft labels from teachers towards student models in both pre-training and fine-tuning phrases.
    \item \textbf{MiniLM}~\citep{minilm}, which distills the self-attention module of the last Transformer layer of the teacher towards the student models.
\end{itemize} 

\paragraph{Structured Pruning}

\begin{itemize}
    \item \textbf{First-order} pruning~\citep{taylor}, which uses the first-order Taylor expansion of the loss towards  attention heads or feed forward neurons to calculate the importance scores.
    \item \textbf{Top-drop}~\citep{toplayer}, which finds it effective to directly discard the top layers of the BERT and fine-tune on the downstream tasks.
    \item \textbf{SNIP}~\citep{snip}, which prunes modules based on the absolute magnitude of their activation outputs, and introduces a spectral normalization mechanism to stabilize the distribution of the post-activation values of the Transformer layers.
    \item \textbf{schuBERT}~\citep{schubert}, which parameterizes each layer of BERT by five different dimensions, and  then pre-train BERT with pruning-based criterions to remove the rows or columns of the parameter matrices of BERT.
\end{itemize} 

\paragraph{Unstructured Pruning} 

\begin{itemize}
    \item \textbf{Magnitude} pruning~\cite{magnitude}, which removes parameters simply accroding to their absolute values.
    \item \textbf{L$_0$-regularization}~\citep{l0}, which sparsifies the model using hard-concrete distribution to exert L$_0$ regularization.
    \item \textbf{Movement} pruning~\citep{movement}, which calculates importance score using Eq.~\ref{eq:movement} in the main content of the paper, and maintains the Top-K parameters.
    \item \textbf{Soft-movement} pruning~\citep{movement}, which is also based on Eq.~\ref{eq:movement} but maintain parameters according to a threshold.
\end{itemize}

\end{document}